\documentclass[10pt,twocolumn,letterpaper]{article}

\usepackage{iccv}
\usepackage{times}
\usepackage{epsfig}
\usepackage{graphicx}
\usepackage{amsmath}
\usepackage{amssymb}
\usepackage{subfigure}
\usepackage{slashbox}
\usepackage{graphicx}
\usepackage{amsmath,amssymb} 
\usepackage{multirow}
\newcommand{\x} {\textbf{x}}

\newcommand{\g} {\textbf{g}}

\newcommand{\y} {\textbf{y}}
\newcommand{\w} {\textbf{w}}
\newcommand{\W} {\textbf{W}}

\newcommand{\I} {\textbf{I}}
\newcommand{\K} {\textbf{K}}

\newcommand{\kk} {\textbf{k}}

\usepackage{algorithmic}

\DeclareMathAlphabet{\mathpzc}{OT1}{pzc}{m}{it} %

\newcommand{\trans}[1]{{#1}^{\ensuremath{\mathsf{T}}}} 

\usepackage[pagebackref=false,breaklinks=true,letterpaper=true,colorlinks,bookmarks=false]{hyperref}

\iccvfinalcopy 

\def\iccvPaperID{458} 

\ificcvfinal\fi
\begin{document}
	
	\title{Learning Social Relation Traits from Face Images}
	
	\author{Zhanpeng Zhang, Ping Luo, Chen Change Loy, and Xiaoou Tang\\
Department of Information Engineering, The Chinese University of Hong Kong\\
		\tt\small zz013@ie.cuhk.edu.hk, pluo@ie.cuhk.edu.hk, ccloy@ie.cuhk.edu.hk, xtang@ie.cuhk.edu.hk
		}
	
	\maketitle

\begin{abstract}
	Social relation defines the association, \eg, warm, friendliness, and dominance, between two or more people.
	Motivated by psychological studies, we investigate if such fine-grained and high-level relation traits can be characterised and quantified from face images in the wild.
	To address this challenging problem we propose a deep model that learns a rich face representation to capture gender, expression, head pose, and age-related attributes, and then performs pairwise-face reasoning for relation prediction.
	To learn from heterogeneous attribute sources, we formulate a new network architecture with a bridging layer to leverage the inherent correspondences among these datasets. It can also cope with missing target attribute labels.
	Extensive experiments show that our approach is effective for fine-grained social relation learning in images and videos. 
	
\end{abstract}

\section{Introduction}
\label{sec:introduction}

\begin{figure}[t]
	\centering
	\includegraphics[width=\linewidth]{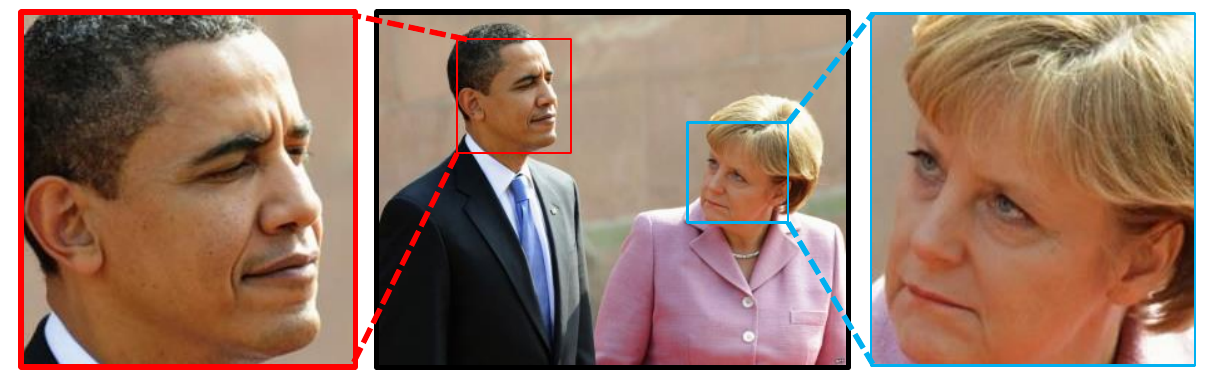}
	\caption{The image is given a caption `\textit{German Chancellor Angela Merkel and U.S. President Barack Obama inspect a military honor guard in Baden-Baden on April 3}.' (source: www.rferl.org). Nevertheless, when we examine the face images jointly, we could observe far more rich social facts that are different from that expressed in the text.}
	\label{fig:introduction}
\end{figure}

Social relation manifests when we establish, reciprocate, or deepen relationships with one another in either physical or virtual world. Studies have shown that implicit social relations can be discovered from texts and microblogs~\cite{fairclough2003analysing}. Images and videos are becoming the mainstream medium to share information, which capture individuals with different social connections. Effectively exploiting such socially-rich sources can provide social facts other than the conventional medium like text (Fig.~\ref{fig:introduction}).

The aim of this study is to characterise and quantify social relation traits from computer vision point of view.
Inspired by extensive psychological studies~\cite{Girard:2014,gottman2001facial,hess2000influence,Knutsonfd}, which show that face emotional expressions can serve as social predictive functions, we wish to automatically recognise fine-grained and high-level social relation traits (\eg,~friendliness, warm, and dominance) from face images.
Such a capability promises a wide spectrum of applications. For instance, automatic social relation inference allows for relation mining from image collection in social network, personal album, and films.


Profiling unscripted social relation from face images is non-trivial. Among the most significant challenges are: (1) as suggested by psychological studies~\cite{Girard:2014,gottman2001facial,hess2000influence}, relations of face images are related to high-level facial factors. Thus we need a rich face representation that captures various attributes such as expression and head pose; (2) no single dataset is presently available, which encompasses all the required facial attribute annotations to learn such a rich representation. In particular, some datasets only contain face expression labels, whilst other datasets may only contain the gender label. Moreover, these datasets are collected from different environments and exhibit different statistical distributions. How to effectively train a model on such heterogeneous data remains an open problem.

\begin{table*}[t]
	\newcommand{\tabincell}[2]{\begin{tabular}{@{}#1@{}}#2\end{tabular}}
	\caption{Descriptions of social relation traits based on~\cite{kiesler1983circle}.}
	\scriptsize
	\label{tab:social_relation_def}
	\setlength{\tabcolsep}{.35667em}
	\begin{center}
		\begin{tabular}{l|c|c}
			\hline
			Relation Trait & Descriptions & Example Pair\\
			\hline
			\hline
			Dominant& one leads, directs, or controls the other / dominates the conversation / gives advices to the other & teacher \& student\\
			\hline
			Competitive& hard and unsmiling / contest for advancement in power, fame, or wealth & people in a debate\\
			\hline
			Trusting& sincerely look at each other / no frowning or showing doubtful expression / not-on-guard about harm from each other &partners\\
			\hline
			Warm& speak in a gentle way / look relaxed / readily to show tender feelings & mother \& baby\\
			\hline
			Friendly& work or act together / express sunny face / act in a polite way / be helpful &host \& guest\\
			\hline
			Attached& engaged in physical interaction / involved with each other / not being alone or separated & lovers\\
			\hline
			Demonstrative& talk freely being unreserved in speech /  readily to express the thoughts instead of keep silent / act emotionally&friends in a party\\
			\hline
			Assured& express to each other a feeling of bright and positive self-concept, instead of depressed or helpless& teammates\\
			\hline
		\end{tabular}
	\end{center}
\end{table*}

To this end, we carefully formulate a deep model to learn a face representation for social relation prediction, driven by rich facial attributes such as expression, head pose, gender, and age. We devise a new deep architecture that is capable of (1) dealing with missing attribute labels from different datasets, and (2) bridging the gap of heterogeneous datasets by weak constraints derived from the association of face part appearances. This allows the model to learn more effectively from heterogeneous datasets with different annotations and statistical distributions.
Unlike existing face analyses that mostly consider single subject, our network is formulated with a Siamese-like architecture~\cite{bromley1994signature}, it is thus capable of jointly considering pairwise faces for relation reasoning, where each face serves as the mutual context to the other.
%

The \textbf{contributions} of this study are three-fold:
(1) to our knowledge, this is the first work that investigates face-driven social relation inference, of which the relation traits are defined based on psychological study~\cite{kiesler1983circle}. We carefully investigate the detectability and quantification of such traits from a pair of face images.
(2) we carefully construct a new social relation dataset labeled with pairwise relation traits supported by psychological studies~\cite{kiesler1983circle,Knutsonfd}, which can facilitate future research on high-level face interpretation.
(3) we formulate a new deep architecture for learning face representation driven by multiple tasks, bridging the gap from heterogeneous sources with potentially missing target attribute labels. It is also demonstrated that the model can be extended to utilize additional cues such as the faces' relative location, besides face images.

\begin{figure*}[t]
	\centering
	\includegraphics[width=1\linewidth]{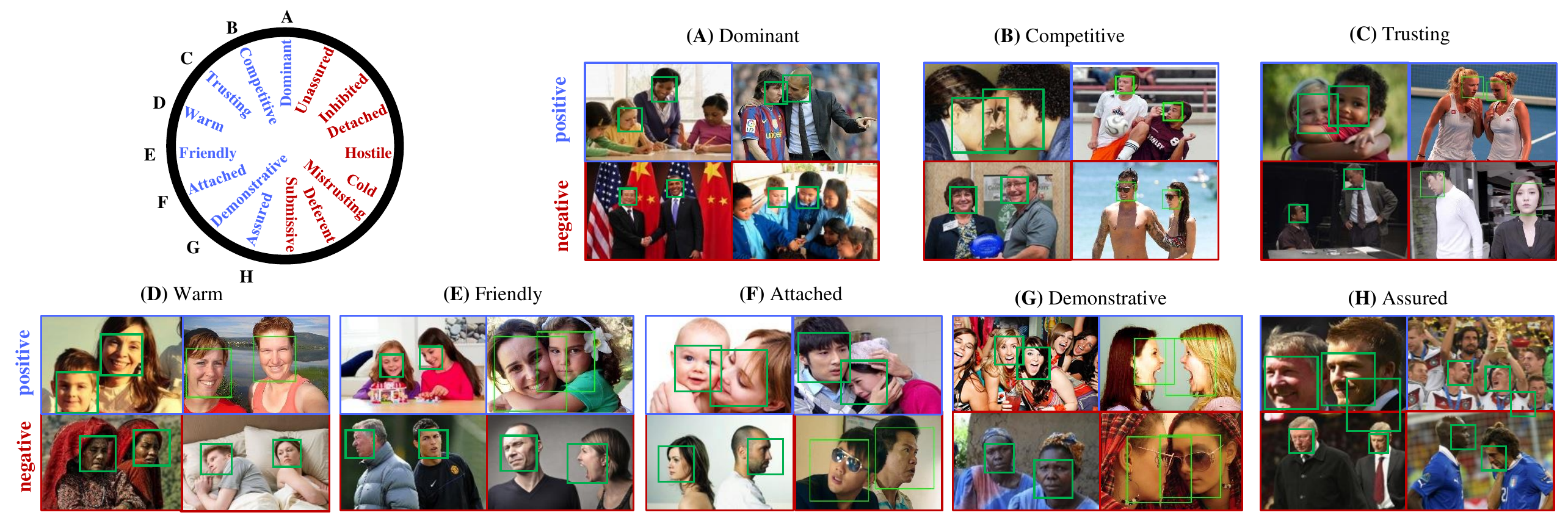}

	\caption{The 1982 Interpersonal Circle (upper left) is proposed by Donald J. Kiesle, and commonly used in psychological studies~\cite{kiesler1983circle}. The 16 segments in the circle can be grouped into 8 relation traits. The traits are non-exclusive therefore can co-occur in an image. In this study, we investigate the detectability and quantification of these traits from computer vision point of view. (A)-(H) illustrate positive and negative examples of the eight relation traits. More detailed definition can be found in the supplementary material.}
	\label{fig:relation_definition}
\end{figure*}

\section{Related Work}
\label{sec:related_work}

\noindent \textbf{Social signal processing.}
Understanding social relation is an important research topic in \textit{social signal processing}~\cite{cristani2013human,pantic2011social,pentland2007social,vinciarelli2009social,vinciarelli2012bridging}, an important multidisciplinary problem that has attracted a surge of interest from computer vision community.
Social signal processing mainly involves facial expression recognition~\cite{liu2013facial} and affective behaviour analysis~\cite{nicolaou2012dynamic}.
On the other hand, there exists a number of studies that aim to infer social relation from images and videos~\cite{ding2010learning,ding2011inferring,fathi2012social,ramanathan2013social,4757440}.
Many of these studies focus on the coarser level of social connection other than the one defined by Kiesler in the interpersonal circle~\cite{kiesler1983circle}. For instance, Ding and Yilmaz~\cite{ding2010learning} only discover social group without inferring relation between individuals.
Fathi \etal~\cite{fathi2012social} only detect three social interaction classes, \ie,~ `dialogue, monologue and discussion'.
Wang \etal~\cite{wang2010seeing} define social relation by several social roles, such as `father-child' and `husband-wife'.
%
%
Other related problems also include image communicative intents prediction~\cite{6909429} and social role inference~\cite{lan2012social}, usually applied on news and talks shows~\cite{raducanu2012inferring}, or meetings to infer dominance~\cite{hung2007using}.

Our work differs significantly from the aforementioned studies.
Firstly, most affective analysis approaches are based on single person therefore cannot be directly employed for interpersonal relation inference.
In addition, these studies mostly focus on recognizing prototypical expressions (happy, angry, sad, disgust, surprise, fear). Social relation is far more complex involving many factors such as age and gender. Thus, we need to consider more attributes jointly in our problem.
Secondly, in comparison to the existing social relation studies~\cite{ding2010learning,fathi2012social}, our work aims to recognize fine-grained and high-level social relation traits~\cite{kiesler1983circle}.
%
Thirdly, many of the social relation studies did not use face images directly for relation inference, but visual concepts~\cite{ding2011inferring} discovered by detectors or people spatial proximity in 2D or 3D space~\cite{chen2012discovering}. All these information sources are valuable for learning human interactions but social relation is fundamentally limited by the input sources.

\vspace{0.1cm}
\noindent \textbf{Human interaction and group behavior analysis.}
%
%
Existing group behavior studies~\cite{hoaitalking,humaninteraction} mainly recognize action-oriented behaviors such as hugging, handshaking or walking, but not social relations. Often, group spatial configuration and actions are exploited for the recognition. Our study differs in that we aim to recognize abstract relation traits from faces.

\vspace{0.1cm}

\noindent \textbf{Deep learning.}
Deep learning has achieved remarkable success in many tasks of face analysis, \eg~face parsing \cite{pluo2}, face landmark detection \cite{zhang}, face attribute prediction \cite{ziwei,pluo1}, and face recognition \cite{sun2013hybrid,pluo3}.
%
However, deep learning has not yet been adopted for face-driven social relation mining that requires joint reasoning from multiple subjects. In this work, we propose a deep model to cope with complex facial attributes from heterogeneous datasets, and joint learning from face pair.

\section{Social Relation Prediction from Face Images}

\subsection{Definitions of Social Relation Traits}
\label{sec:relation_traits}


We define the social relation traits based on the interpersonal circle proposed by Kiesler~\cite{kiesler1983circle}, where human relations are divided into 16 segments as shown in Fig.~\ref{fig:relation_definition}.
Each segment has its opposite side in the circle, such as ``friendly and hostile''. Therefore, the 16 segments can be considered as eight binary relations, whose descriptions and examples are given in Table~\ref{tab:social_relation_def}. More detailed descriptions are provided in the \textit{supplementary material}. We also provide positive and negative visual samples for each relation in Fig.~\ref{fig:relation_definition}, showing that they are visually perceptible.
For instance, ``friendly'' and ``competitive'' are easily separable because of the conflicting meanings. However, some relations are close such as ``friendly'' and ``trusting'', implying that a pair of faces can have more than one social relation.
%


\subsection{Social Relation Dataset}
\label{sec:social_relation_dataset}
To investigate the detectability of social relations from a pair of face images, we build a new dataset\footnote{\url{http://mmlab.ie.cuhk.edu.hk/projects/socialrelation/index.html}}, containing $8,306$ images chosen from web and movies. Each image is labelled with faces' bounding boxes and their pairwise relations. This is the first face dataset measuring social relation traits and it is challenging because of large face variations including poses, occlusions, and illuminations.

%
%
%

We carefully built this dataset.
Five performing arts students were asked to label each relation for each face image independently. Thus, each label has five annotations. A label is accepted if more than three annotations are consistent. The inconsistent samples were presented again to the five annotators to seek consensus\footnote{The average Fleiss' kappa of the eight relation traits' annotation is 0.62, indicating substantial inter-rater agreement.}.
%
%
%
To facilitate the annotation task, we also provide multiple cues to the annotators. First, to help them understand the social relations, we list ten related adjectives defined by ~\cite{kiesler1983circle} for the positive and negative samples on each relation trait, respectively. Multiple example images are also provided. Second, for the image frames selected from the movies, the annotators were asked to get familiar with the stories. The subtitles were presented during labelling.

%
%


\begin{figure*}[!t]
	\centering
	
	\includegraphics[width=1.0\textwidth]{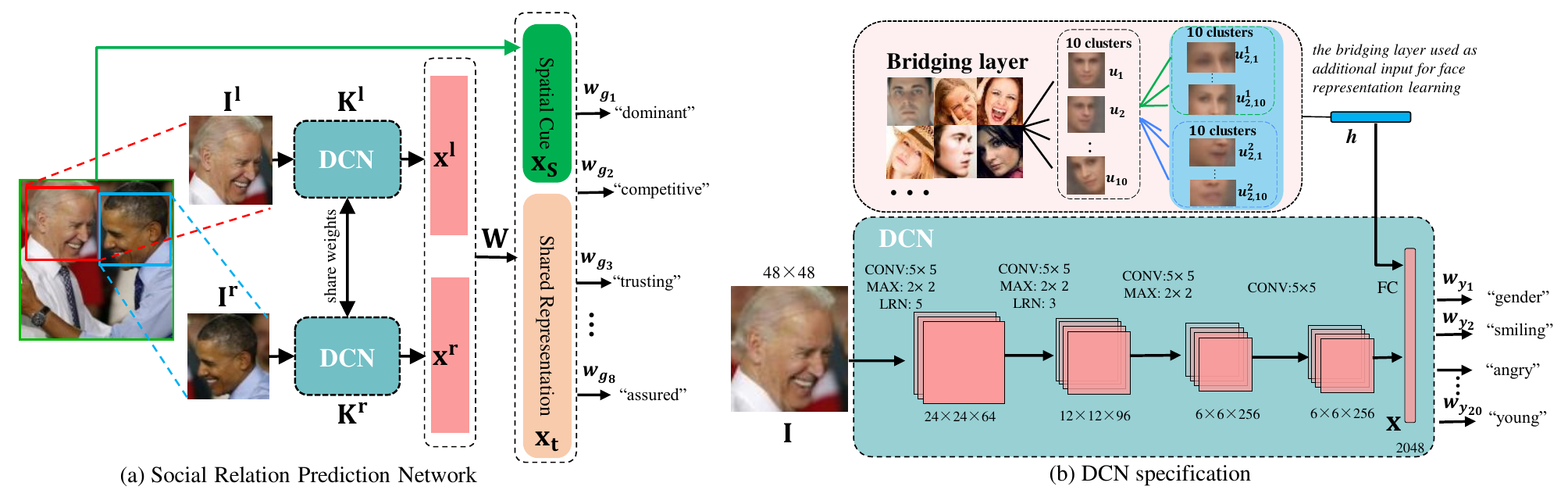}

	\caption{(a) Overview of the network for interpersonal relation learning. (b) The new deep architecture we propose to learn a rich face representation driven by sematic attributes. This network is used as the initialization for the DCN in (a) for relation learning. The operation of ``CONV'', ``MAX'', ``LRN'' and ``FC'' denote convolution, max-pooling, local response normalization and fully-connected, respectively. The numbers following the operations are the parameters for kernel size.}
	\label{fig:pipeline}
\vspace{-0.2cm}
\end{figure*}


\subsection{Baseline Method}
\label{sec:baseline_method}

To predict social relations from face images, we first introduce a strong baseline method by using a Siamese-like deep convolutional network (DCN), which learns an end-to-end mapping from raw pixels of a pair of face images to relation traits.
DCN is effective for learning shared representations as demonstrated in \cite{taigman2014deepface}.
As shown in Fig.\ref{fig:pipeline}(a), given an image of social relation, we detect a pair of face images, denoted as $\I^r$ and $\I^l$, from which we extract high-level features $\x^r$ and $\x^l$ using two DCNs respectively, $\forall\x^r,\x^l\in\mathbb{R}^{2048\times1}$.
These two DCNs have identical network structures, where
$\K^r$ and $\K^l$ denote the network parameters, which are tied to increase generalization ability.
A weight matrix, $\W\in\mathbb{R}^{4096\times256}$, projects the concatenated feature vectors to a space of shared representation $\x_t$, which is utilised to predict a set of relation traits, $\g=\{g_i\}_{i=1}^8$, $\forall g_i\in\{0,1\}$. Each relation is modeled as a single binary classification task, parameterized by a weight vector, $\w_{g_i}\in\mathbb{R}^{256\times1}$.

To improve the baseline method, we incorporate some spatial cues to train the deep network as shown in Fig.\ref{fig:pipeline}(a), which includes 1) two faces' positions $\{x^l,y^l,w^l,h^l,x^r,y^r,w^r,h^r\}$, representing the $x$-,$y$-coordinates of the upper-left corner, width, and height of the bounding boxes; $w^l$ and $w^r$ are normalized by the image width. Similar for $h^{l}$ and $h^r$; 2) the relative faces' positions: $\frac{x^l-x^r}{w^l},\frac{y^l-y^r}{h^l}$, and 3) the ratio between the faces' scales: $\frac{w^l}{w^r}$. The above spatial cues are concatenated as a vector, $\x_s$, and combined with the shared representation $\x_t$ for learning relation traits.

As the above description, each binary variable $g_i$ can be predicted by linear regression,
\begin{equation}
g_i=\trans{\w}_{g_i}[\x_s;\x_t]+\epsilon,
\end{equation}
where $\epsilon$ is an additive error random variable, which is distributed following a standard logistic distribution, $\epsilon\sim Logistic(0,1)$. $[\cdot;\cdot]$ indicates the column-wise concatenation of two vectors.
Therefore, the probability of $g_i$ given $\x_t$ and $\x_s$ can be written as a sigmoid function, $p(g_i=1|\x_t,\x_s)=1/(1+\exp\{-\trans{\w}_{g_i}[\x_s;\x_t]\})$,
indicating that $p(g_i|\x_t,\x_s)$ is a Bernoulli distribution,
$p(g_i|\x_t,\x_s)=p(g_i=1|\x_t,\x_s)^{g_i}
\big(1-p(g_i=1|\x_t,\x_s)\big)^{1-g_i}$.

In addition, the probabilities of $\w_{g_i}$, $\W$, $\K^l$, and $\K^r$ can be modeled by the standard normal distributions. For example, suppose $\K$ contains $K$ filters, then $p(\K)=\prod_{j=1}^Kp(\kk_j)=\prod_{j=1}^K\mathcal{N}(\textbf{0},\mathcal{I})$, where $\textbf{0}$ and $\mathcal{I}$ are an all-zero vector and an identity matrix respectively, implying that the $K$ filters are independent. Similarly, we have $p(\w_{g_i})=\mathcal{N}(\textbf{0},\mathcal{I})$.
Furthermore, $\W$ can be initialized by a standard matrix normal distribution \cite{gupta1999matrix}, \ie $p(\W)\propto\exp\{-\frac{1}{2}\mathrm{tr}(\W\trans{\W})\}$, where $\mathrm{tr}(\cdot)$ indicates the trace of a matrix.
%
%

Combining the above probabilistic definitions, the deep network is trained by maximising a posterior probability,
\begin{equation}\label{eq:MAP}
\begin{split}
\arg&\max_{\Omega}~p(\{\w_{g_i}\}_{i=1}^8,\W,\K^l,\K^r|\g,\x_t,\x_s,\I^r,\I^l)\propto\\
&\Big(\prod_{i=1}^8 p(g_i|\x_t,\x_s)p(\w_{g_i})\Big)\Big(\prod_{j=1}^K p(\kk^l_j)p(\kk^r_j)\Big)p(\W),\\
&~~~~~~~~~~~~~~~~~~\mathrm{s.t.}~~~~~~\K^r=\K^l
\end{split}\vspace{-10pt}
\end{equation}
where $\Omega=\{\{\w_{g_i}\}_{i=1}^8,\W,\K^l,\K^r\}$ and the constraint means the filters are tied.
Note that $\x_t$ and $\x_s$ represent the hidden features and the spatial cues extracted from the left and right face images, respectively. Thus, the variable $g_i$ is independent with $\I^l$ and $\I^r$, given $\x_t$ and $\x_s$.

By taking the negative logarithm of Eqn.(\ref{eq:MAP}), it is equivalent to minimising the following loss function
\begin{equation}\label{eq:E}
\begin{split}
&\arg\min_\Omega\sum_{i=1}^8\Big\{\trans{\w}_{g_i}\w_{g_i}-(1-g_i)\ln\big(1-p(g_i=1|\x_t,\x_s)\big)
-\\
&g_i\ln p(g_i=1|\x_t,\x_s)\Big\}+\sum_{j=1}^K(\trans{\kk^r_j}\kk^r_j
+\trans{\kk^l_j}\kk^l_j)+\mathrm{tr}(\W\trans{\W}),\\
&~~~~~~~~~~~~~~~~~~\mathrm{s.t.}~~~~~~\kk_j^r=\kk_j^l,~j=1...K
\end{split}
\end{equation}
where the second and the third terms correspond to the traditional cross-entropy loss, while the remaining terms indicate the weight decays~\cite{moody1995simple} of the parameters. Eqn.(\ref{eq:E}) is defined over single training sample and is a highly nonlinear function because of the hidden features $\x_t$. It can be efficiently solved by stochastic gradient descent~\cite{krizhevsky2012imagenet}.

\subsection{A Cross-Dataset Approach}
\label{sec:cross_dataset_network}

As investigated by the psychological studies~\cite{Girard:2014,gottman2001facial,hess2000influence}, the social relations of face images are strongly related to some hidden high-level factors, such as emotion. Learning these semantic concepts implicitly from raw image pixels imposes great challenge. To explicitly learn these factors, an ideal solution is to introduce two additional loss functions on top of $\x^l$ and $\x^r$ respectively, representing that not only the concatenation of $\x^l$ and $\x^r$ learns the relation traits, but each of them also learns the high-level factors of its corresponding face image. However, this solution is impractical, because labelling both social relations and emotions of face images is too expensive.

To overcome this limitation, we extend the baseline model by pre-training the DCN with face attributes, which are borrowed from existing face databases. These attributes capture the high-level factors, guiding the predictions of relation traits. The advantages are three folds: 1) face attributes, such as age, gender, and expressions, are highly correlated with the high-level factors of social relations, as supported by the psychological studies~\cite{Girard:2014,gottman2001facial,hess2000influence,Knutsonfd}; 2) leveraging the existing face databases not only improves generalized capacity but also make data preparation much easier; and 3) the face representation induced by semantic attributes can bridge the gap between the high-level relation traits and low-level image pixels.



\begin{table*}[t]
	\newcommand{\tabincell}[2]{\begin{tabular}{@{}#1@{}}#2\end{tabular}}
	\caption{Summary for the labelled attributes in the datasets: AFLW~\cite{6130513}, CelebFaces~\cite{sun2013hybrid} and Kaggle Expression~\cite{Goodfeli-et-al-2013}.}
	\footnotesize
	\label{tab:attribute_datasets}
	\begin{center}
\setlength{\tabcolsep}{.6667em}
		\begin{tabular}{l|c|c|c|c|c|c|c|c|c|c|c|c|c|c|c|c|c|c|c|c}
			\hline
			 \multirow{2}[10]{*}{Attributes}&Gender&\multicolumn{5}{c|}{Pose}&\multicolumn{9}{c|}{Expression}&\multicolumn{5}{c}{Age}\\\cline{2-21}
			&\rotatebox{90}{gender}&\rotatebox{90}{left profile}&\rotatebox{90}{left}&\rotatebox{90}{frontal}&\rotatebox{90}{right}&
			\rotatebox{90}{right profile}&
			\rotatebox{90}{angry}&\rotatebox{90}{disgust}&\rotatebox{90}{fear}&
			 \rotatebox{90}{happy}&\rotatebox{90}{sad}&\rotatebox{90}{surprise}&\rotatebox{90}{neutral}&\rotatebox{90}{smiling}&
			\rotatebox{90}{\parbox{1.1cm}{mouth \\ opened}}&
			\rotatebox{90}{young}&\rotatebox{90}{goatee}&
			\rotatebox{90}{no beard}&\rotatebox{90}{sideburns}&
			\rotatebox{90}{\parbox{1.5cm}{5 o'clock \\ shadow}}\\
			\hline\hline
			AFLW&$\surd$&$\surd$&$\surd$&$\surd$&$\surd$&$\surd$&&&&&&&&&&&&&\\
			
			 CelebFaces&$\surd$&$\surd$&$\surd$&$\surd$&$\surd$&$\surd$&&&&&&&&$\surd$&$\surd$&$\surd$&$\surd$&$\surd$&$\surd$&$\surd$\\
			
			 \tabincell{l}{Kaggle}&&&&&&&$\surd$&$\surd$&$\surd$&$\surd$&$\surd$&$\surd$&$\surd$&&&&&&\\
			\hline
		\end{tabular}
	\end{center}
	\vspace{-0.45cm}
\end{table*}
\vspace{0.2cm}

In particular, we make use of data from three public datasets, including AFLW~\cite{6130513}, CelebFaces~\cite{sun2013hybrid}, and Kaggle~\cite{Goodfeli-et-al-2013}. Different datasets have been labelled with different sets of face attributes. A summary is given in Table~\ref{tab:attribute_datasets}, where the attributes are partitioned into four groups.


It is clear that the training datasets are from multiple heterogenous sources and they have been labelled with different sets of attributes. For instance, AFLW only contains gender and poses, while Kaggle only has expressions. In addition, these datasets exhibit different statistical distributions, causing issues during pre-training. It can be shown that if we perform joint training directly, each attribute is trained by the labelled data alone, instead of benefitting from the existence of the unlabelled data.
Consider a simple example of three datasets, denoted as $A$, $B$, and $C$, where $A$ and $B$ are labelled with attribute $y^{1}$ and $y^{2}$ respectively, while dataset $C$ is labelled with $y^{1}$, $y^{2}$ and $y^{3}$. Moreover, $\x_{A}$ indicates a training sample from dataset $A$.
Given three training samples $\x_{A}$, $\x_{B}$, and $\x_{C}$, attribute classification is to maximise the joint probability $p(y^{1}_A,y^{2}_A,y^{3}_A,$ $y^{1}_B,y^{2}_B,y^{3}_B,$
$y^{1}_C,y^{2}_C,y^{3}_C|\x_{A},\x_{B},\x_{C})$.
Since the samples are independent and $A$ and $B$ only contain attributes $y^1$ and $y^2$ respectively, the joint probability can be factorized as $p(y^{1}_{A},y^{2}_A,y^{3}_A|\x_{A})$ $\cdot$ $p(y^{1}_B,y^{2}_B,y^{3}_B|\x_{B})$ $\cdot$ $p(y^{1}_{C},y^{2}_{C},y^{3}_{C}|\x_{C})$ $=$ $p(y^{1}_{A}|\x_{A})$ $\cdot$ $p(y^{2}_B|\x_{B})$ $\cdot$ $p(y^{1}_{C},y^{2}_{C},y^{3}_{C}|\x_{C})$. For example, we have $\sum_{y^{2}_A,y^{3}_A}p(y^{1}_{A},y^{2}_A,y^{3}_A|\x_{A})$ $=$ $p(y^{1}_{A}|\x_{A})$.
As the attributes are also independent, the joint probability can be further written as $p(y^{1}_{A},y^{1}_{C}|\x_{A},\x_{C})p(y^{2}_{B},
y^{2}_{C}|\x_{B},\x_{C})p(y^{3}_{C}|\x_{C})$, indicating that each attribute classifier is trained by the labelled data alone. For instance, the classifier of the first attribute is trained by data from $A$ and $C$.

\noindent\textbf{Bridging the gaps between multiple datasets.} Since faces from different datasets share similar structure in local part, such as mouth and eyes, we propose a bridging layer based on the local correspondence to cope with the different dataset distributions. In particular, we establish a face descriptor $h$ based on the mixture of aligned facial parts.
As shown in Fig.~\ref{fig:pipeline}(b), we build a three-level hierarchy to partition the facial parts' shape, where each child node groups the data of its parents into clusters, such as $u^{1}_{2,1}$ and $u^{1}_{2,10}$. In the top layer, the faces are divided into 10 clusters by K-means using the landmark locations from the SDM face alignment algorithm~\cite{6618919}. Each cluster captures the topological changes due to viewpoints. Fig.~\ref{fig:pipeline}(b) shows the mean face of each cluster. In the second layer, for each node, we perform K-means using the locations of landmarks in the upper and lower face region, and obtain 10 clusters respectively. These clusters captures the local shape of the facial parts. Then the mean HOG feature of the faces in each cluster is regarded as the corresponding template. Given a new sample, the descriptor $h$ is obtained by concatenating its L2-distance to each template.
In this case, the descriptor $h$ serves as a correspondence label for datasets. We use it as additional input in the fully connected layer for facial feature $\x$ (see Fig.\ref{fig:pipeline}(b)). Thus the learned face representations for samples from different datasets are driven to be close if the correspondence labels are similar.
It is worth noting that this bridging layer is different from the work of~\cite{ahmed2008training,weston2012deep}, where the algorithms build some clusters from training data as an auxiliary task. Differently, the proposed method uses the aligned facial part association, which is well suited for our problem, instead of simply construct the cluster from the whole image. Moreover, since the construction of $h$ is unsupervised, it contains noise and may harm the training if used as targets. Instead, we use the descriptor as additional input, which shows better performance than used as output (see Table.~\ref{tab:attribute_accuracy}). The rest of the DCN structure is described in Fig.\ref{fig:pipeline}(b), which includes four convolutional layers, three max-pooling layers, two local response normalization layers, and two fully-connected layers. The rectified linear unit~\cite{krizhevsky2012imagenet} is adopted as the activation function.


Then the DCN objective is to predict a set of attributes $\y=\{y_l\}_{l=1}^{20}$, $\forall y_l\in\{0,1\}$. Each relation is modeled as a single binary classification task, parameterized by a weight vector, $\w_{y_l}\in\mathbb{R}^{2048\times1}$. The probability of $y_l$ can be computed by a sigmoid function. Similar to Eqn.(\ref{eq:E}), it can be formulated as minimising the cross-entropy loss.

\noindent\textbf{Learning procedure.} Similar to the relation prediction network, the training process can be done by back-propagation (BP) using stochastic gradient descent (SGD)~\cite{krizhevsky2012imagenet}. The difference is that we have missing attribute labels in the training set.
Specifically, we use the cross-entropy loss for attribute classification, with an estimated attribute $\widetilde{y}_l$, the back-propagation error $e^l$ is
\begin{equation}\label{eq:error1}
	e^{t}=
	\begin{cases}
		0  & \text{if $y_l$ is missing,}\\
		y_l-\widetilde{y}_l & \text{otherwise.}
	\end{cases}
\end{equation}


\section{Experiments}

\noindent \textbf{Facial attribute datasets.}
\label{sec:facial_attribute_datasets}
To enable accurate social relation prediction, we employ three datasets to cover a wide-range of facial attributes: Annotated Facial Landmarks in the Wild (AFLW)~\cite{6130513} (24,386 faces), CelebFaces~\cite{sun2013hybrid} (87,628 faces) and a facial expression dataset on Kaggle contest~\cite{Goodfeli-et-al-2013} (35,887 faces).
Table~\ref{tab:attribute_datasets} summarises the data.
%
All the attributes are binary and labelled manually. To evaluate the performance of the cross dataset approach, we randomly select 2,000 testing faces from AFLW and CelebFaces, respectively. For the Kaggle dataset, we follow the protocol of the expression contest by using the 7,178 testing faces.

\vspace{0.2cm}
\noindent \textbf{Social relation dataset.}
\label{sec:social_relation_dataset2}
%
%
%
We build the social relation dataset as described in Sec.~\ref{sec:social_relation_dataset}. Table~\ref{tab:relation_datasets} presents the statistics of this dataset.
%
%
Specially, to reduce the potential effect from annotators' subjectivity, we select a subset (522 cases) from the testing images and build an additional testing set. The images in this subset are all from movies. As the annotators know the movies' story, they can give objective annotation assisted by the subtitle.

\begin{table}[t]
	\caption{Statistics of the social relation dataset.}
	\footnotesize
	\label{tab:relation_datasets}
	\begin{center}
		\setlength{\tabcolsep}{.58667em}
		\begin{tabular}{l|c|c|c|c}
			\hline
			\multirow{2}{*}{Relation trait}&\multicolumn{2}{c|}{training}&\multicolumn{2}{c}{testing}\\\cline{2-5}
			&\#positive&\#negative&\#positive&\#negative\\
			\hline \hline
			dominant&418&7041&112&735\\
			
			competitive&538&6921&123&724\\
			
			trusting&6288&1171&609&238\\
			
			warm&6224&1235&619&228\\
			
			friendly&6790&669&734&113\\
			
			attached&6407&1052&695&152\\
			
			demonstrative&6555&904&699&148\\
			
			assured&6595&864&685&162\\
			\hline
		\end{tabular}
	\end{center}
	\vspace{-0.5cm}
\end{table}

\subsection{Social Relation Trait Prediction}
\label{sec:Predicting_Interpersonal_Relation}

\begin{figure*}[t]
	\centering
	\includegraphics[width=1\textwidth]{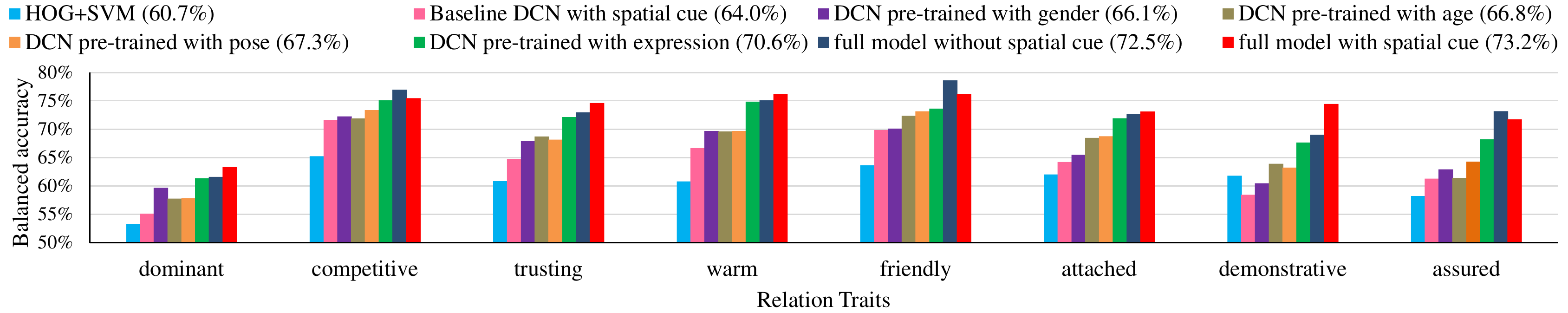}
	\caption{Relation traits prediction performance. The number in the legend indicates the average accuracy of the according method across all the relation traits.}
	\label{fig:performance}
\end{figure*}

\begin{figure*}
	\centering
	\includegraphics[width=1\textwidth]{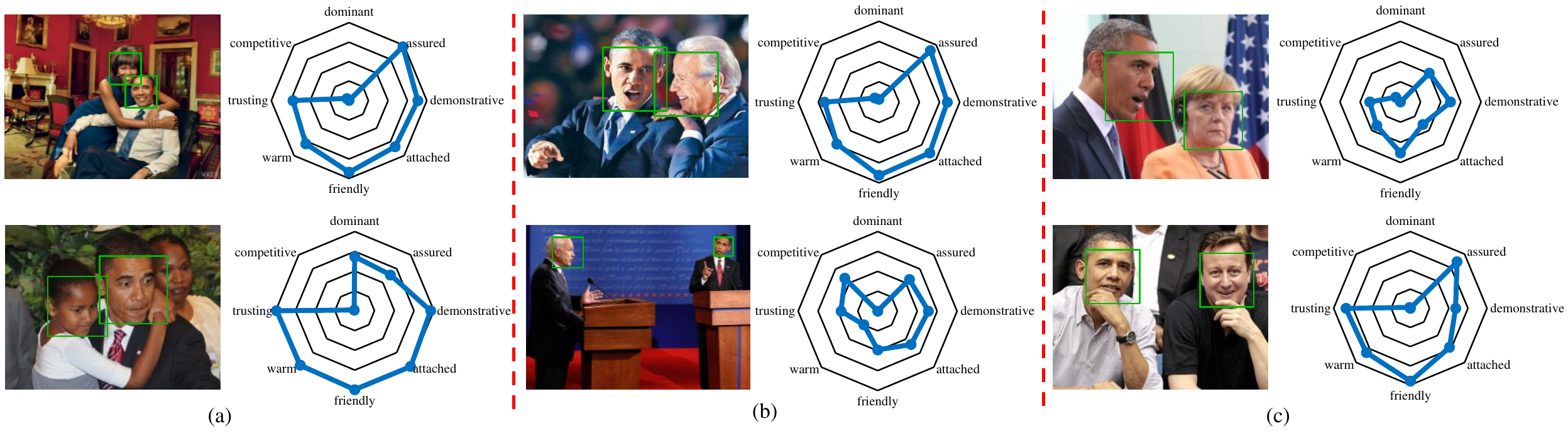}
	\vskip -0.1cm
	\caption{The relation traits predicted by our full model with spatial cue. The polar graph beside each image indicates the tendency for each trait to be positive.}
	\label{fig:samples}
	\vspace{-0.4cm}
\end{figure*}

\noindent
\textbf{Baseline algorithm.} In addition to the strong baseline method in Sec.~\ref{sec:baseline_method}, we train an additional baseline classifier by extracting the HOG features from the given face images. The features from the two faces are then concatenated and we use a linear support vector machine (SVM) to train a binary classifier for each relation trait. For simplicity, we call this method ``HOG+SVM'', and the baseline method in Sec.~\ref{sec:baseline_method} ``Baseline DCN''.

\vspace{0.1cm}
\noindent
\textbf{Performance evaluation.} We divide the relation dataset into training and testing partitions of 7,459 and 847 images, respectively. The face pairs in these two partitions are mutually exclusive. To account for the imbalance positive and negative samples, a balanced accuracy is adopted:
\begin{equation}\label{eq:evaluation}
accuracy=0.5(n_p/N_p+n_n/N_n),
\end{equation}
where $N_p$ and $N_n$ are the numbers of positive and negative samples, whilst $n_p$ and $n_n$ are the numbers of true positive and true negative.
We first train the network as Sec.~\ref{sec:baseline_method} (\ie, Baseline DCN). After that, to examine the influences of different attribute groups, we pre-train four DCN variants using only one group of attribute (expression, age, gender, and pose). In addition, we compare the effectiveness between the full model with and without spatial cue.

Fig.~\ref{fig:performance} shows the accuracies of the different variants.
All variants of our deep model outperform the baseline HOG+SVM.
We observe that the cross dataset pre-training is beneficial, since pre-training with any of the attribute groups improves the overall performance.
In particular, pre-training with expression attributes outperforms other groups of attributes (improving from 64.0\% to 70.6\%). This is not surprising since social relation is largely manifested from expression.
The pose attributes come next in terms of influence to relation prediction.
The result is also expected since when people are in a close or friendly relation, they tend to look at the same direction or face each other.
Finally, the spatial cue is shown to be useful for relation prediction.
%
%
However, we also observe that not every trait is improved by the spatial cue and some are degraded. This is because currently we simply use the face scale and location directly, of which the distribution is inconsistent in images from different sources. As for the relation traits, ``dominant'' is the most difficult trait to predict as it needs to be determined by more complicated factors, such as the social role and environmental context. The trait of ``assured'' is also difficult since it is visually subtle compared to other traits such as ``competitive'' and ``friendly''.
In addition, we conduct analysis on the movie testing subset. Table~\ref{tab:movie_subset} shows the average accuracy on the eight relation traits of the two baseline algorithms and the proposed method. The results correspond to that of the whole testing set. This supports the reliability of the proposed dataset.

\begin{table}[t]
	\centering
	\caption{Balanced accuracies (\%) on the movie testing subset.}
	\vspace{0.1cm}
	\label{tab:movie_subset}
	\setlength{\tabcolsep}{.20667em}
	\begin{tabular}{c|c|c|c}
		
		\hline  Method&HOG+SVM&\parbox{2.3cm}{Baseline DCN \\with spatial cue}&  \parbox{2.3cm}{Full model\\with spatial cue}\\
		\hline
		\hline  Accuracy& 58.92\% & 63.76\% & 72.6\% \\
		\hline
	\end{tabular}
\vspace{-10pt}
\end{table}

\begin{figure}
	\centering
	\includegraphics[width=0.45\textwidth]{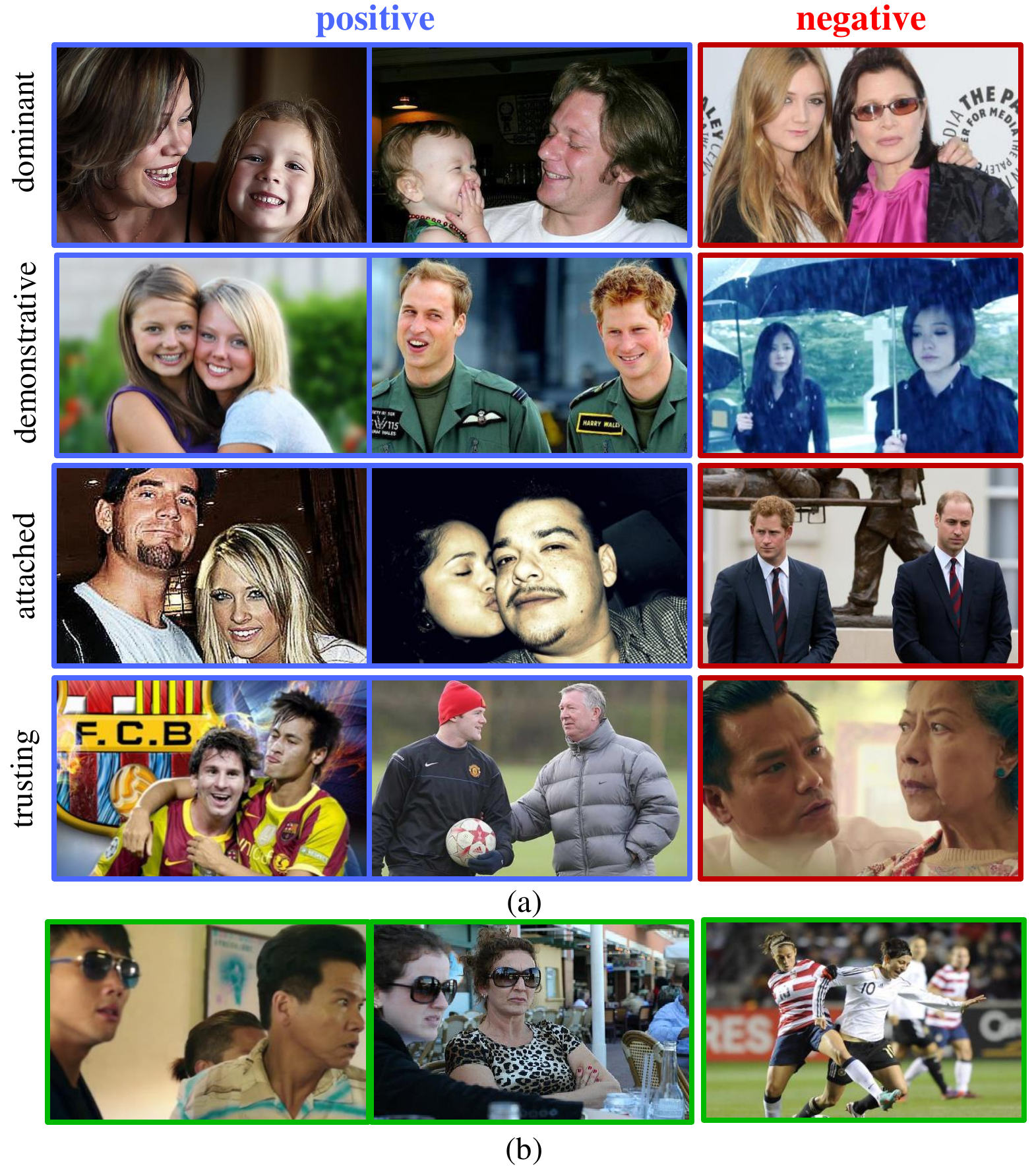}
	\caption{(a) Positive and negative prediction results on different relation traits. (b) False positives on ``assured'', ``demonstrative'' and ``friendly'' relation traits (from left to right).}
	\label{fig:samples2}
\vspace{-0.4cm}
\end{figure}

Some qualitative results are presented in Fig.~\ref{fig:samples}.
Positive relation traits, such as ``trusting'', ``warm'', ``friendly'' are inferred between the US President \textit{Barack Obama} and his family members. Interestingly, ``dominant'' trait is predicted between him and his daughter (Fig.~\ref{fig:samples}(a)).
The upper image in Fig.~\ref{fig:samples}(b) was taken in his election celebration party with the US Vice President \emph{Joe Biden}. We can see the relation is quite different from that of the lower image, in which Obama was in the presidential election debate. Fig.~\ref{fig:samples}(c) includes the images for \emph{Angela Merkel}, Chancellor of Germany and \emph{David Cameron}, Prime Minister of UK. The upper image is usually used in the news articles on US spying scandal, showing low probability on the ``trusting'' trait. More positive and negative results on different relation traits are shown in Fig.~\ref{fig:samples2} (a). In addition, we show some false positives in Fig.~\ref{fig:samples2} (b), which are mainly caused by faces with large occlusions.

\begin{table*}[t]
	\newcommand{\tabincell}[2]{\begin{tabular}{@{}#1@{}}#2\end{tabular}}
	\caption{Balanced accuracies (\%) over different attributes with and without bridging layer (BL).}
    \vspace{0.2cm}
	\footnotesize
	\label{tab:attribute_accuracy}
	\begin{center}
		\setlength{\tabcolsep}{.40667em}
		\begin{tabular}{l|c|c|c|c|c|c|c|c|c|c|c|c|c|c|c|c|c|c|c|c|c c c}
			\hline
			 \multirow{2}[15]{0.3cm}{Attributes}&\multirow{2}[33]{*}{\rotatebox{90}{$\qquad\qquad\;\;\;\;$average}}&Gender&\multicolumn{5}{c|}{Pose}&\multicolumn{9}{c|}{Expression}&\multicolumn{5}{c}{Age}\\\cline{3-22}
			&&\rotatebox{90}{gender}&
			\rotatebox{90}{left profile}&
			\rotatebox{90}{left}&
			\rotatebox{90}{frontal}&
			\rotatebox{90}{right}&
			\rotatebox{90}{right profile}&\rotatebox{90}{angry}&\rotatebox{90}{disgust}&\rotatebox{90}{fear}&
			 \rotatebox{90}{happy}&\rotatebox{90}{sad}&\rotatebox{90}{surprise}&\rotatebox{90}{neutral}&\rotatebox{90}{smiling}
			&\rotatebox{90}{\parbox{1.3cm}{mouth opened}}&
			\rotatebox{90}{young}&\rotatebox{90}{goatee}&\rotatebox{90}{no beard}&\rotatebox{90}{sideburns}&
			\rotatebox{90}{\parbox{1.3cm}{5 o'clock \\ shadow}}\\
			\hline\hline
			 HOG+SVM&72.6&81.2&86.8&{71.7}&88.3&\textbf{74.5}&90.1&61.2&63.7&59.2&77.8&60.2&74.8&66.3&83.2&78.9&67.1&60.8&67.8&70.3&67.2\\
			
			Without BL&78.3&92.4&90.2&69.8&87.8&67.3&88.7&64.5&74.5&55.2&87.9&57.3&80.1&66.7&90.9&92.0&79.1&{77.4}&88.1&76.5&79.3\\

			\tabincell{l}{BL as output}&{81.3}&{92.9}&{91.7}&70.1&{90.0}&70.6&{90.2}&{69.1}&{77.0}&\textbf{64.0}&\textbf{91.0}&\textbf{66.1}&\textbf{86.6}&{73.9}&{91.5}&\textbf{92.4}&\textbf{83.5}&74.5&\textbf{91.2}&{79.5}&\textbf{80.6}\\
\tabincell{l}{BL as input}&\textbf{82.4}&\textbf{93.8}&\textbf{92.2}&\textbf{73.4}&\textbf{95.4}&72.5&\textbf{90.4}&\textbf{69.8}&\textbf{79.4}&{63.3}&{90.9}&65.4&{85.3}&\textbf{74.9}&\textbf{92.8}&91.7&{83.2}&\textbf{82.1}&90.3&\textbf{81.7}&80.0\\
\hline
		\end{tabular}
	\end{center}
\vspace{-10pt}
\end{table*}

\begin{figure*}[t]
	\centering
	\includegraphics[width=1\textwidth]{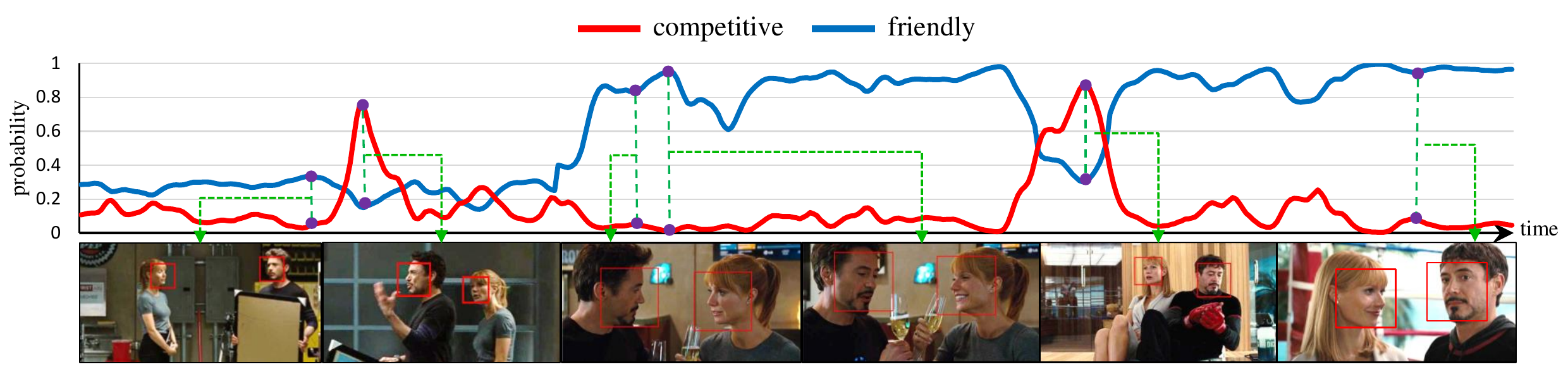}
	\caption{Prediction for relation traits of ``friendly'' and ``competitive''for the movie \emph{Iron Man}. The probability indicates the tendency for the trait to be positive. It shows that the algorithm can capture the friendly talking scene and the moment of confliction.}
	\label{fig:vis_video}
\vspace{-10pt}
\end{figure*}

\subsection{Further Analyses}


\noindent
\textbf{Facial expression recognition.} Given the essential role of expression attributes, we further evaluate our cross dataset approach on the challenging Kaggle facial expression dataset. Following the protocol in~\cite{Goodfeli-et-al-2013}, we classify each face into one of the seven expressions, (\ie~angry, disgust, fear, happy, sad, surprise, and neutral).
%
%
%
The Kaggle winning method~\cite{tang2013deep} reports an accuracy of 71.2\% by applying a CNN with SVM loss function. Our method achieves a better performance of 75.10\%, through fusing data from multiple sources with the proposed bridging layer.


\begin{figure}[t]
	\centering
	\includegraphics[width=0.42\textwidth]{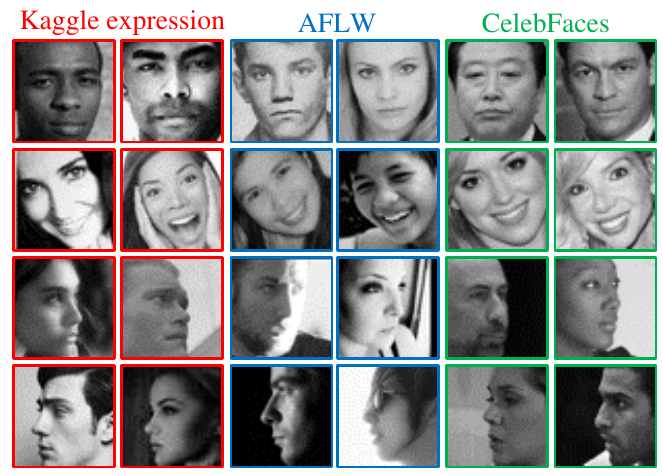}
	\caption{Test samples from different datasets are automatically grouped into coherent clusters by the face descriptor of bridging layer (Sec.~\ref{sec:cross_dataset_network}). Each row corresponds to a cluster.}
	\label{fig:vis_cluster}
\end{figure}

\vspace{0.4cm}
\noindent
\textbf{The effectiveness of bridging layer.}
We examine the effectiveness of the bridging layer from two perspectives. First, we show some clusters discovered by using the face descriptor (Sec.~\ref{sec:cross_dataset_network}). It is observed that the proposed approach successfully divides samples from different datasets into coherent clusters of similar face patterns.
%
%
Second, we examine the balanced accuracy (Eqn.~(\ref{eq:evaluation})) of attribute classification with and without the bridging layer (Table~\ref{tab:attribute_accuracy}). 
%
%
It is observed that bridging layer benefits the recognition of most attributes, especially the expression attributes. The results suggest the bringing layer an effective way to combine heterogeneous datasets for visual learning by deep network. Moreover, treating bridging layer as input provides higher accuracy than as output.
\subsection{Application: Character Relation Profiling}

%
We show an example of application on using our method to profile the relations among the characters in a movie automatically.
Here we choose the movie \emph{Iron Man}. We focus on different interaction patterns, such as conversation and conflict, of the main roles ``\emph{Tony Stark}'' and ``\emph{Pepper Potts}''. Firstly, we apply a face detector to the movie and select the frames capturing the two roles. Then, we apply our algorithm on each frame to infer their relation traits. The predicted probabilities are averaged across 5 neighbouring frames to obtain a smooth profile. Fig.~\ref{fig:vis_video} shows a video segment with the traits of ``friendly'' and ``competitive''. Our method accurately captures the friendly talking scene and the moment when Tony and Pepper were in a conflict (where the ``competitive'' trait is assigned with a high probability while the ``friendly'' trait is low).

\section{Conclusion}
In this paper we investigate a new problem of predicting social relation traits from face images. This problem is challenging in that accurate prediction relies on recognition of complex facial attributes. We have shown that deep model with bridging layer is essential to exploit multiple datasets with potential missing attribute labels. Future work will integrate face cues with other information such as environment context and body gesture for relation prediction. We will also investigate other interesting applications such as relation mining from image collection in social network. Moreover, we can also explore modelling relations of more than two people, which can be implemented by voting or graphical model, where each node is a face and edge is relations between faces.


{\footnotesize
\bibliographystyle{ieee}
\bibliography{short,cavan_bib,zp_bib}
}

\end{document}